\documentclass[11pt,letterpaper]{article}

\usepackage[margin=1in]{geometry}
\usepackage[T1]{fontenc}
\usepackage{newtxtext,newtxmath}
\usepackage{microtype}

\usepackage{graphicx}
\usepackage[figuresright]{rotating}
\usepackage{tabularx}
\usepackage{booktabs}
\usepackage{caption}
\usepackage{subcaption}

\usepackage{authblk}
\usepackage[numbers,sort&compress]{natbib}

\usepackage[
    colorlinks=true,
    linkcolor=blue,
    citecolor=blue,
    urlcolor=blue
]{hyperref}

\AtBeginDocument{%
  \setlength{\abovedisplayskip}{6pt}%
  \setlength{\belowdisplayskip}{6pt}%
  \setlength{\abovedisplayshortskip}{6pt}%
  \setlength{\belowdisplayshortskip}{6pt}%
}

\title{Predicting Surface Displacement Fields of
Laser Powder Bed Fusion Parts Using STL-Derived
Graphs and Part-Level Descriptors}

\author[1]{Saima Ahmed Suhi}
\author[1]{Lang Yuan}
\author[1]{Md Habibor Rahman\thanks{
Corresponding author:
\href{mailto:habiborrahman@sc.edu}
{habiborrahman@sc.edu}}}

\affil[1]{\small {Department of Mechanical Engineering, University of South Carolina,
Columbia, SC 29208, USA}}

\date{}

\begin{document}

\maketitle

\begin{abstract}
Predicting part distortion before fabrication remains a key challenge in the Laser Powder Bed Fusion (LPBF) process. Thermomechanical simulation is commonly used to predict such distortion by modeling thermal and mechanical responses, but it can be computationally demanding for repeated design evaluation. In response, this work predicts the three-dimensional surface displacement field of LPBF parts directly from their STL derived graphs employing a two-branch Graph Attention Network (GAT). One branch combines normalized geometric information from connected vertices of a densified STL surface mesh using learned attention weights; the other encodes 12 part-level descriptors characterizing mass distribution, geometric contact area with the build plate, overhang, size, and shape. Training labels are transferred from displacement fields simulated using ANSYS ADDITIVE PRINT onto the STL surface vertices through a Gaussian-kernel weighted average. The GAT model was evaluated on 117 parts with a 5-fold cross-validation split by parts, achieving an $R^2$ of $0.946$ and a MAE error of $0.0178$ mm for displacement magnitude. Compared with the node-wise MLP, GAT increases $R^2$ from 0.913 to 0.946, with the largest component-wise improvement occurring in the build direction, where $U_z$ $R^2$ increases from 0.903 to 0.939. GAT also achieves higher predictive accuracy than GraphSAGE (displacement-magnitude $R^2$ = 0.936) and Graph Convolutional Network (GCN) (displacement-magnitude $R^2$ = 0.934) in all five folds. Additionally, an ablation study evaluates the contribution of each descriptor group by removing it individually and retraining the model. Removing the mass-distribution descriptors reduces displacement-magnitude $R^2$ from 0.946 to 0.900, while removing descriptors of geometric contact with the build plate and overhang reduces it to 0.920. These comparisons indicate that combining information from connected surface vertices with part-level geometric descriptors improves accuracy. By linking CAD geometry directly to surface nodal displacement with graphs and global descriptors, the approach supports faster design iteration, earlier identification of distortion-prone features, and more informed LPBF production planning.
\end{abstract}

\vspace{0.5em}

\section{Introduction}
\label{Introduction}

Laser powder bed fusion (LPBF) enables the fabrication of metal components with complex geometries, but maintaining dimensional accuracy remains a major challenge \cite{enayet2026data}. Repeated localized heating and cooling during the build process produce nonuniform thermal expansion and contraction, leading to residual stress and part distortion \cite{Chen2026a, Vulimiri2025}. Predicting distortion before printing enables early assessment of part geometries and helps guide design modifications to improve dimensional accuracy. \enlargethispage{-\baselineskip} Distortion is commonly predicted using finite-element simulations, including thermomechanical models and simplified approaches based on inherent or assumed strains \cite{Dong2024,Qi2025}. These simulations capture the mechanisms governing residual stress and distortion within parts, but running a simulation for each part can take hours to days, depending on geometric complexity, mesh resolution, and the modeling approach \cite{Huang2026}. In practice, designers compare many variants, orientations, and support design strategies before printing, so running a full simulation for every candidate slows the design iteration and is often deferred until later in the design process \cite{Tao2025}. A faster approach that predicts distortion directly from the design geometry would allow candidate geometries to be screened before starting the manufacturing process. Data-driven surrogate models offer an alternative to repeated numerical simulation of thermal and mechanical responses in Metal Additive Manufacturing (MAM). Some existing approaches use geometric encodings and process parameters to predict part-level quantities or categories, such as peak deviation, feature recognition, and geometry clusters \cite{Aljarrah2023,wang2023hybrid}. Although useful for assessing overall part behavior or grouping geometries, these outputs cannot indicate the distortion region.

Beyond part-level predictions, studies are increasingly using geometric representations and process parameters to learn spatially resolved responses \cite{Li2026, Mozaffar2021}. Under fixed material, process, and boundary conditions, part geometry influences distortion through its effect on heat transfer, temperature evolution, and mechanical constraint \cite{gao2024experimental,bai2026layer}. During LPBF part fabrication, the surrounding material and build-plate attachment restrict thermal contraction, while the part shape determines available heat-dissipation paths and the stiffness resisting deformation \cite{wang2026orientation,Makeen2024}. The resulting displacement field therefore depends on how material is distributed along the build direction, how much of the part is attached to the plate, and how much of its surface is unsupported or down-facing. These are properties of the geometry itself, suggesting that a model with an adequate geometric information can learn spatially resolved distortion without explicit thermal inputs. Researchers have used voxel grids, signed distance fields, and point clouds for this purpose \cite{Saimon2026,Dong2022}, but voxelization trades resolution for computational cost \cite{Shonkwiler2025} and produces stair-stepped surfaces, while point clouds do not explicitly encode surface connectivity.

Triangulated surface meshes, by contrast, are the native format of design files (e.g., STL) and explicitly encode connectivity between surface vertices. These meshes can be converted into graphs by treating surface vertices as nodes and mesh edges as connections between them, thereby preserving the spatial relationships among neighboring vertices \cite{Wong2022,Wu2021}. By aggregating information from connected vertices, graph-based models can capture spatial dependencies relevant to predicting temperature and displacement fields across a part. For example, Mozaffar et al. (2021) used a graph-based framework to predict thermal histories \cite{Mozaffar2021} in Directed Energy Deposition (DED), while subsequent studies by Raut et al. (2025) and Choi et al. (2025) investigated the transferability of graph-based thermal prediction in LPBF and DED, respectively \cite{Raut2025,Choi2025}. These studies demonstrate how mesh connectivity, spatial coordinates, and information about heat-source locations and boundary conditions can be utilized to predict temperature changes over time at different locations within a part. Graph-based modeling has also been extended to spatial prediction of residual stress and distortion. For example, Li et al. (2026) developed a Graph Attention Network (GAT) to predict three-dimensional residual stress in DED workpieces using nodal coordinates and temperature information \cite{Li2026}. Chen et al. (2026) introduced a DeepONet-GNN framework that integrates sparse temperature histories with structural features, including spatial coordinates, boundary conditions, and initial displacements, to predict distortion along the build direction \cite{Chen2026}. Lee et al. (2025) developed GraphCompNet, which uses point-cloud representations and incorporates part-position information to predict and compensate for manufacturing-induced shape deviations \cite{lee2025graphcompnet}. Collectively, these studies demonstrate the potential of graph-based models to capture spatial relationships and incorporate structural and process-related information for distortion prediction. In a related but distinct application, Giovenali et al. (2026) introduced PAGE-Net, which integrates local mesh features, global geometric descriptors, and printing parameters to estimate manufacturing resource requirements in fused filament fabrication \cite{Giovenali2026}. Although developed for a different prediction task, PAGE-Net demonstrates the feasibility of combining local mesh representations with global geometric descriptors within a graph-based learning framework.

Previous studies have manifested the potential of graph-based models for predicting manufacturing-induced distortion and integrating local and global geometric information, but the combined use of these representations to predict full-field LPBF displacement directly from part geometry remains underexplored. Such an approach could enable early assessment of geometric distortion without requiring thermal histories, process-state measurements, or additional simulations for individual parts at the prediction stage. To bridge this gap, this study proposes a two-branch Graph Attention Network (GAT) that combines local surface features and connectivity with twelve part-level geometric descriptors to predict displacement at each surface vertex, without requiring simulation results as inputs during inference. The study develops an STL-based surrogate for predicting three-dimensional surface displacement fields of LPBF parts under specific material and process conditions. A densified STL surface provides both the prediction graph and the locations at which simulation-derived displacement labels are assigned. Gaussian-kernel weighted transfers displacement vectors from the corresponding VTK\footnote{VTK stands for Visualization Toolkit. In this study, the VTK files store the simulation mesh and the true displacement vector at each node.} simulation surface to the STL vertices, bridging the different design and simulation discretizations. The contribution extends beyond the architectural combination by assessing the label-transfer procedure and controlling the predictive inputs. Held-out reconstruction tests assess label transfer accuracy, while comparisons with a capacity-matched node-wise MLP and alternative graph operators examine the value of message passing. Additionally, a descriptor drop-group ablation retraining identifies the complementary information supplied by part-level geometry descriptors. Together, these analyses establish and evaluate a geometry-informed workflow for predicting LPBF surface displacement from STLs, supporting its potential use in design-stage distortion screening. The primary objectives of this study are the following:

\begin{enumerate}
    \item Develop a two-branch graph attention network framework to predict full-field three-dimensional surface displacement in LPBF directly from STL-derived mesh geometry and geometric descriptors under fixed material and process conditions, enabling design-stage distortion assessment without requiring additional process simulations for individual predictions.

    \item Construct and validate a simulation-to-STL displacement-transfer workflow using surface densification and Gaussian-kernel-weighted averaging to generate reliable training labels on the design mesh.

    \item Evaluate the effectiveness of graph-based feature aggregation and the contribution of part-level geometric descriptors to prediction accuracy through controlled model comparisons and descriptor-group ablation studies.

\end{enumerate}
\begin{figure*}[htb]
  \centering
  \includegraphics[width=.9\textwidth]{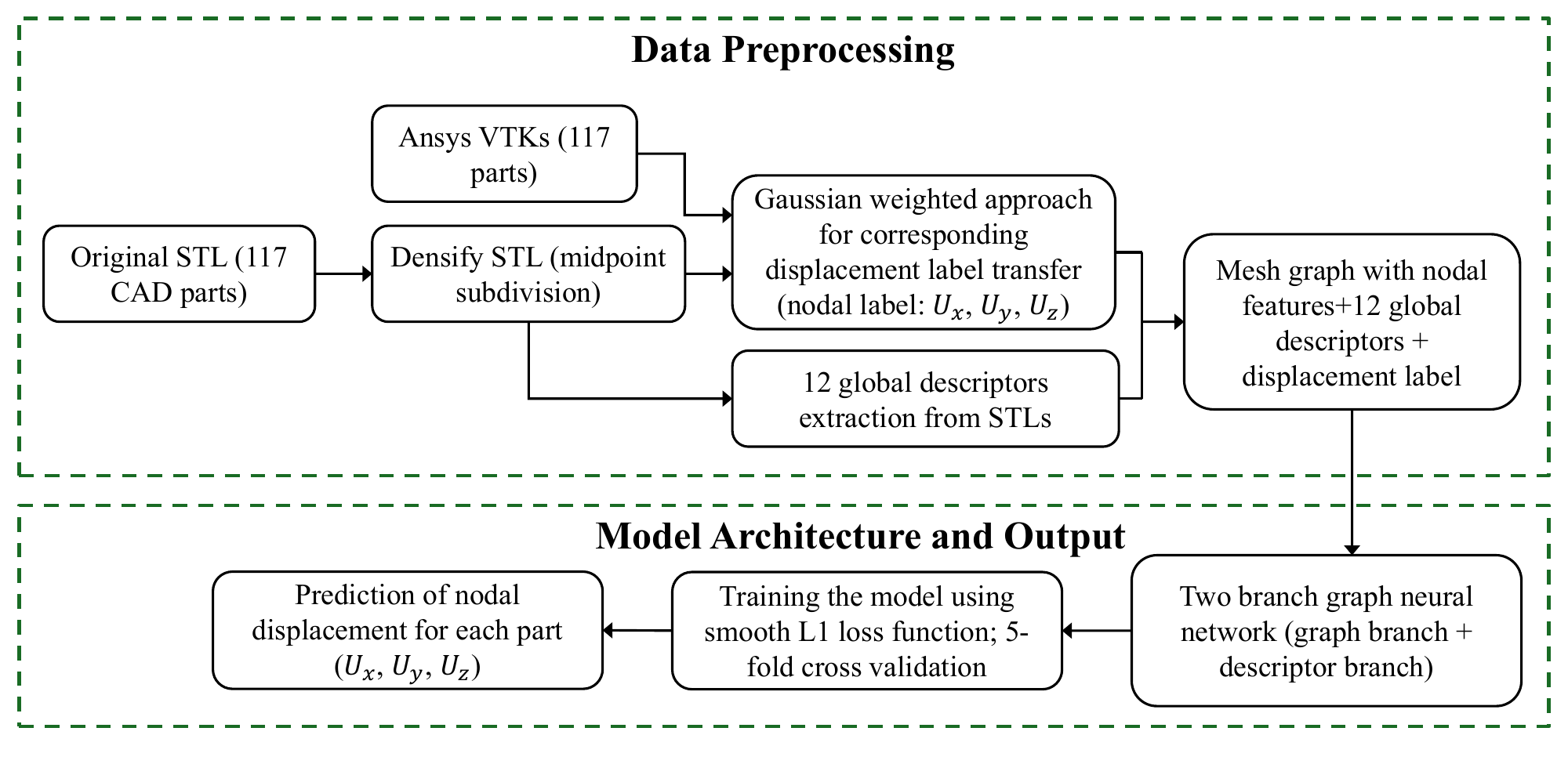}
  \caption{Overview of the pipeline: STL densification, Gaussian-kernel label transfer from the simulation, descriptor extraction, and the two-branch graph neural network.}
  \label{fig:outline_paper}
\end{figure*}

\section{Methodology}
\label{Methodology}
This section describes how the training data were generated and how the model was built, trained, and evaluated as illustrated in Fig.~\ref{fig:outline_paper}. The process simulation is used only offline to supply the displacement labels. Subsection~\ref{Dataset} describes the dataset; Subsections~\ref{Data pre-processing}--\ref{Part-level global descriptors} describe how each STL–simulation pair becomes a labeled surface graph with part-level descriptors; Subsections~\ref{Model architecture}--\ref{Evaluation Metrics} present the network, its training, and the evaluation protocol.

\subsection{Dataset}
\label{Dataset}
\begin{figure}[t]
  \centering
  \includegraphics[width=.6\columnwidth]{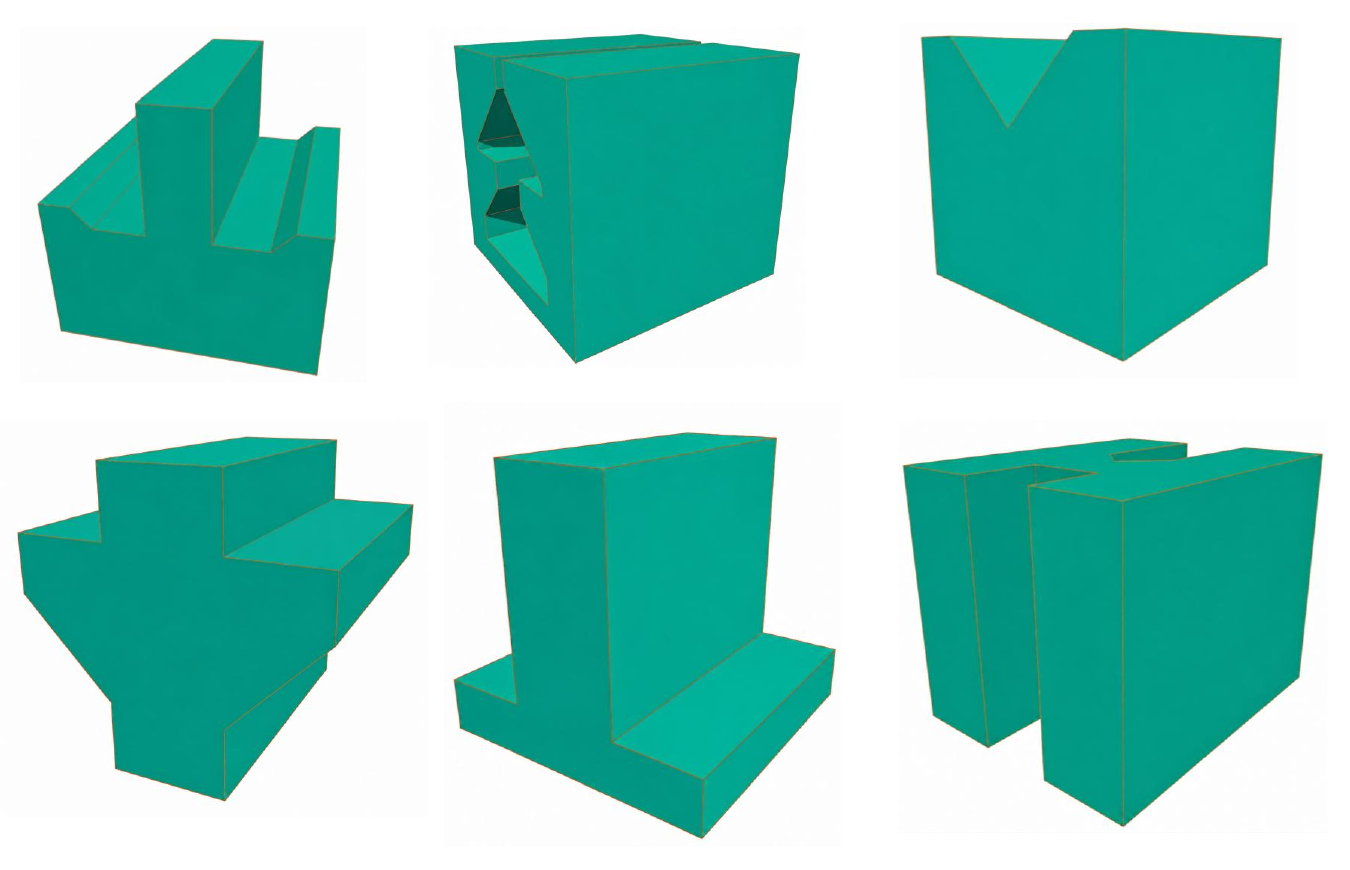}
  \caption{Six example parts from the dataset of 117 LPBF parts, showing the range of
  geometric features: slots, steps, inclined faces and internal cavities.}
  \label{fig:dataset}
\end{figure}

This study uses an extension of the LPBF simulation dataset reported in \cite{Bushra2025, bushra2023}. It comprises 117 parts with features such as slots, steps, and chamfers (Fig.~\ref{fig:dataset}), each provided as an STL file with the corresponding Ansys process simulation result. All parts were simulated with a single process condition: the same material, process parameters, and baseplate constraint. Geometries are expressed in the build frame, with $z$ as the build direction. Bounding-box dimensions ranges from 15--39~mm, volumes range from 3{,}565 to 51{,}422~mm$^3$, and the peak displacement magnitude ranges from 0.12 to 1.28~mm (median is around 0.50~mm). 

\subsection{Data pre-processing}
\label{Data pre-processing}
The simulation encodes 3D displacements at mesh nodes, which generally do not coincide with the STL vertices, because the simulation uses a much denser mesh. Training labels were therefore constructed on the STL surface. Each STL mesh was first densified through triangle subdivision because its original tessellation was too sparse to adequately resolve spatial variations in displacement. Simulated displacements were then transferred to the vertices of the densified mesh with a Gaussian-weighted average (Nadaraya--Watson) \cite{nadaraya1964, watson1964}, and vertices with unreliable transferred labels were masked during training and evaluation (about 0.04\% of the total surface nodes). The following subsections describe these steps.

\subsubsection{STL densification}
\label{STL densification}
Design STL files can represent planar faces using a few large triangles, leaving too few vertices to adequately resolve spatial variations in displacement. Each mesh was therefore refined by recursive midpoint subdivision using (\texttt{trimesh}, \texttt{subdivide\_to\_size}) until no edge exceeded \(1.5~\mathrm{mm}\), and duplicate vertices were merged. Because midpoint subdivision only splits existing triangles, every new vertex lies on the original triangle, preserving the represented surface geometry. Surface area, volume, and bounding-box bounds were compared with the original mesh using a relative tolerance of $10^{-6}$ both before and after export and all the meshed parts match their original meshes. The resulting meshes have approximately 1.8k--19.8k vertices per part. The maximum edge length was chosen as a compromise between resolving the displacement field and limiting the graph size, and hence the training time. All 117 meshes are watertight and consistently oriented, with no boundary or non-manifold edges. Densification increases sampling density without adding geometric detail; curved surfaces retain the faceting of the source STL.

\subsubsection{Label transfer from simulation to the STL surface}
\label{Label transfer from simulation to the STL surface}
The boundary surface of each simulation mesh was extracted in undeformed coordinates, together with its nodal displacement vectors. No rotation or scaling was applied during the transfer.

The displacement at each STL vertex $\mathbf{x}_i$ was computed as a Gaussian kernel weighted (Nadaraya--Watson) average of nearby simulation surface nodes $\mathbf{x}_j$ \cite{nadaraya1964,watson1964,siraskar2024application} (for noncoincident STL vertices):
\begin{equation}
  \mathbf{u}_i=\frac{\sum_{j\in\mathcal{S}_i} w_{ij}\,\mathbf{u}_j}{\sum_{j\in\mathcal{S}_i} w_{ij}},
  \qquad
  w_{ij}=\exp\!\left(-\frac{\lVert\mathbf{x}_i-\mathbf{x}_j\rVert^2}{2\sigma^2}\right)
  \label{eq:transfer}
\end{equation}
Here, the neighbor set $\mathcal{S}_i$ was restricted to the
connected surface component that contains the simulation node nearest to $\mathbf{x}_i$, so that disconnected surface shells were not mixed. However, the dataset didn't contain any such points; it was kept as a safety check during the Gaussian operation. The spherical Gaussian kernel bandwidth was $\sigma=0.375$~mm, corresponding to \(1.5\) voxels. Candidate simulation surface nodes were searched within an initial radius of 0.5~mm, which was enlarged by a factor of 1.5, up to 3~mm, until it contained at least 6 nodes, and at most the 12 nearest nodes were used. For 97.8\% of vertices, the initial radius was sufficient; the largest required radius was 1.125~mm, with an average of 11.7 neighbors per vertex. STL vertices that coincide with a simulation surface node (within $10^{-8}D$; 2.1\% of all vertices) take the displacement of that node directly.

\subsubsection{Label validation and masking}
\label{Label validation and masking}
A held-out reconstruction test was conducted to assess the Gaussian kernel weighted average independently of model training. For each of 10 selected parts, 10\% of the simulation surface nodes were randomly withheld, and their displacement vectors were reconstructed using only the remaining 90\%, with the same Gaussian bandwidth, adaptive search radii, neighbor limits, and connected-component restriction used for label transfer. Held-out nodes were excluded from all source queries, and no exact-node shortcut was used. Three random repeats were performed per part. All held-out nodes were successfully reconstructed, and the mean vector MAE across the 30 trials was 0.00228~mm. This test evaluates label transfer accuracy at simulation surface nodes; it does not directly quantify errors arising from geometric mismatch between the STL and simulation surfaces.

In case of masking, an STL vertex was marked invalid when its distance to the closest point on the simulation surface exceeded 0.5\% of the part's bounding-box diagonal (0.17--0.33~mm, depending on part size), or when fewer than 6 neighbors were found within the maximum radius of 3~mm. At such vertices, the voxel surface does not follow the STL surface closely enough for the transferred label to be trusted. Invalid vertices remain in the graph and participate in message passing, but are excluded from the loss and all metrics. Across the dataset, only 592 of about 1.4 million vertices (0.04\%) in 11 parts were masked, all by the distance criterion; the largest fraction was in p2080, which is 12.2\% of the part's vertices. 

\subsubsection{Surface graph and node features}
\label{Surface graph and node features}
Each densified mesh becomes an undirected graph $\mathcal{G}=(\mathcal{V},\mathcal{E})$ whose nodes are the mesh vertices and whose edges are the triangle edges, stored bidirectionally. Each node carries the 10 features explained in Table~\ref{tab:nodefeat}. 

Let $\mathbf{x}_i=(x_i,y_i,z_i)$ be the position of vertex $i$,
$\mathbf{c}$ the bounding-box centre, $D$ the bounding-box diagonal,
$H$ the part height, $z_{\min}$ the lowest $z$ coordinate,
$\mathbf{n}_i$ the outward unit vertex normal computed using
\texttt{trimesh} and re-normalized to unit length,
$\mathcal{N}(i)$ the vertices connected to node $i$ by a mesh edge,
and $A_f$ the area of triangle $f$. The notation $f\ni i$ denotes
triangles incident to vertex $i$, $a_i$ is the surface area assigned
to vertex $i$, and $\bar{a}$ is its mean over all vertices of the part.

\begin{table}[t]
  \centering
  \caption{Node features (10 per vertex).}
  \label{tab:nodefeat}

  \setlength{\tabcolsep}{2pt}
  \renewcommand{\arraystretch}{1.5}
  \begin{tabularx}{\columnwidth}{@{}lXc@{}}
    \toprule
    Feature & Definition & Dim. \\
    
    \midrule
   
    Normalized position & $(\mathbf{x}_i-\mathbf{c})/D$ & 3 \\
    Relative build height & $(z_i-z_{\min})/H$ & 1 \\
    Vertex normal & $\mathbf{n}_i$ & 3 \\
    Normal variation & $\kappa_i=\frac{1}{|\mathcal{N}(i)|}\sum_{j\in\mathcal{N}(i)}\left(1-\mathbf{n}_i\!\cdot\!\mathbf{n}_j\right)$ & 1 \\
    Relative vertex area & $a_i/\bar{a}$, \; $a_i=\sum_{f\ni i}A_f/3$ & 1 \\
    Radial distance & $\lVert\mathbf{x}_i-\mathbf{c}\rVert/D$ & 1 \\
    
    \bottomrule

  \end{tabularx} 
\end{table}

The normal-variation term $\kappa_i$ is zero where neighboring vertex normals are identical and increases with local normal variation. All coordinates are divided by the same diagonal $D$, preserving the part's aspect ratio.

\subsubsection{Part-level global descriptors}
\label{Part-level global descriptors}
Twelve part-level descriptors are computed from the STLs (Table~\ref{tab:descriptors}), and none of them uses simulation output.

Let $L_x, L_y, L_z$ be the bounding-box dimensions, $S$ the total surface area, $V$ the enclosed volume, and $z_c$ the $z$-coordinate of the volume centroid. Base faces were identified using proximity to the lowest build plane, with a tolerance of $\delta_b=\max(10^{-4}H,10^{-6}~\mathrm{mm})$, and an outward normal within $5^\circ$ of the negative $z$-direction. The set of base faces is denoted by $\mathcal{F}_{\mathrm{base}}$. $A_{\text{base}}^{xy}$ denotes their summed area projected onto the $xy$-plane; the resulting base-footprint fraction is clipped
to $[0,1]$ to accommodate numerical round-off. The overhang and down-facing area fractions include only non-base faces and are normalized by $S$. These faces are classified using the $z$-component of their outward unit normal, $n_{z,f}$: overhang faces satisfy $n_{z,f}\leq-\cos45^\circ+10^{-6}$, where $10^{-6}$ is a numerical tolerance, and down-facing faces satisfy $n_{z,f}\leq-\sin0.1^\circ$.

\begin{table}[t]
  \centering
  \caption{Global geometric descriptors (12 per part) and the groups used in the descriptor ablation.}
  \label{tab:descriptors}

  \setlength{\tabcolsep}{2pt}
  \renewcommand{\arraystretch}{1.5}
  \begin{tabular*}{\columnwidth}{@{\extracolsep{\fill}}lll@{}}   
    \toprule
    Group & Descriptor & Definition \\
    \midrule
    Size & Bounding box & $L_x,\;L_y,\;L_z$ \\
         & Surface area & $S$ \\
         & Volume & $V$ \\
         & Height/width & $L_z/\max(L_x,L_y)$ \\
    \addlinespace
    Surface-to-volume ratio & Surface-to-volume ratio & $S/V$ \\
    \addlinespace
    Mass distribution & Fill ratio & $V/(L_xL_yL_z)$ \\
                      & Centroid height & $(z_c-z_{\min})/L_z$ \\
    \addlinespace
    Base contact and overhang & Base footprint & $A_{\text{base}}^{xy}/(L_xL_y)$ \\
                 & Overhang ($45^\circ$) & $\displaystyle \sum_{\substack{f\notin\mathcal{F}_{\mathrm{base}}\\ n_{z,f}\le-\cos45^\circ+10^{-6}}} A_f/S$ \\
                 & Down-facing 
                 & $\displaystyle \sum_{\substack{f\notin\mathcal{F}_{\mathrm{base}}\\
                 n_{z,f}\le-\sin0.1^\circ}} A_f/S$ \\
    \bottomrule
  \end{tabular*}
    
\end{table}

The surface-to-volume ratio $S/V$ has units of inverse length and, in some cases, provides a geometric interpretation of wall thickness, although it mostly reflects overall part size and shape. The fill ratio and normalized centroid height characterize the spatial distribution of the part volume within its bounding box. The base-footprint descriptor characterizes geometric contact with the lowest build plane. In contrast, the overhang and down-facing descriptors quantify the fractions of non-base surface area facing downward under the specified angular criteria.

\subsection{Model architecture}
\label{Model architecture}
A graph attention network \cite{velickovic2018gat} is incorporated into a two-branch architecture for distortion prediction. The architecture combines local surface geometry with whole-part properties, including size, base footprint, and the spatial distribution of the part volume. The graph branch passes messages along mesh edges, allowing each vertex to learn from its neighborhood. The descriptor branch encodes the 12 global geometric descriptors into a part-level embedding, providing explicit information about absolute part size alongside the node features. At each vertex, the graph embedding is concatenated with the corresponding part-level embedding and decoded into the three standardized displacement components.

\begin{figure*}[!htb]
  \centering
  \includegraphics[width=\textwidth]{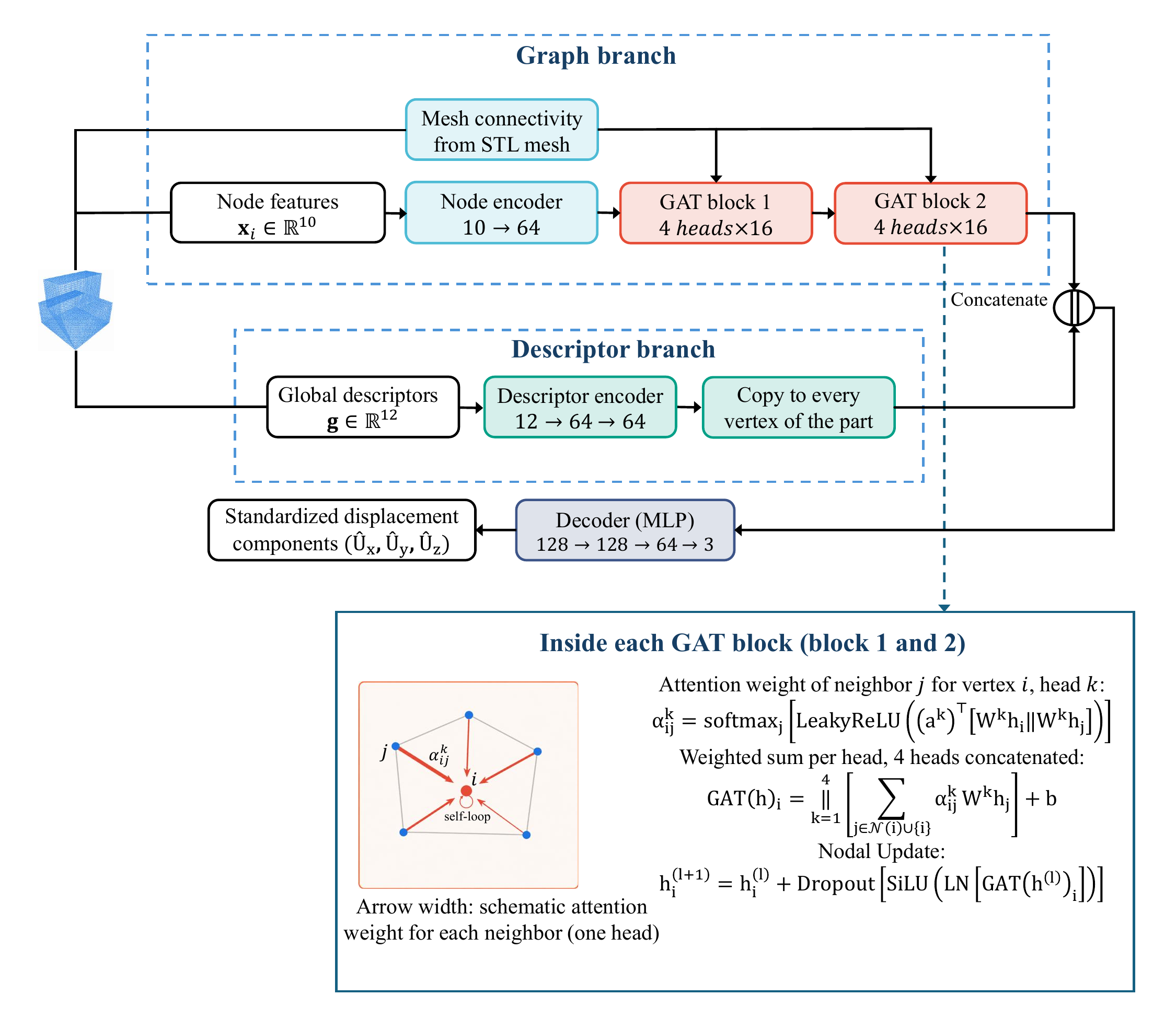}
  \caption{Architecture of the two-branch GAT model. The graph branch encodes the node features and passes them through two residual GAT blocks over the mesh connectivity; the descriptor branch encodes the 12 global descriptors once per part and copies them to every vertex. The model concatenates both and decodes them into standardized displacement components at each vertex. Bottom: operations inside each GAT block, with a schematic of the attention weights of one head.}
  \label{gat}
\end{figure*}

\textbf{Graph branch.} A linear encoder with layer normalization and SiLU activation maps the 10 node features to $d=64$ channels. Two residual message-passing blocks follow, each computing,
\begin{equation}
  \mathbf{h}_i^{(l+1)}=\mathbf{h}_i^{(l)}+\operatorname{Dropout}\ \Big[\operatorname{SiLU}\big(\operatorname{LN}\big(\operatorname{MP}(\mathbf{h}^{(l)})_i\big)\big)\Big]
  \label{eq:block}
\end{equation}
with dropout 0.15. In the proposed model, MP is a graph attention layer (GAT) \cite{velickovic2018gat} with $K=4$ heads of 16 channels, concatenated to 64:
\begin{equation}
  \operatorname{MP}(\mathbf{h})_i=\Big\Vert_{k=1}^{K}\Bigl[\sum_{j\in\mathcal{N}(i)\cup\{i\}}\alpha_{ij}^{k}\,\mathbf{W}^{k}\mathbf{h}_j\Bigr]\;+\;\mathbf{b}
  \label{eq:gat}
\end{equation}
\begin{equation}
  \alpha_{ij}^{k}=\operatorname{softmax}_{j}\ \Big[\operatorname{LeakyReLU}\big(\big(\mathbf{a}^{k})^\top(\mathbf{W}^{k}\mathbf{h}_i\,\Vert\,\mathbf{W}^{k}\mathbf{h}_j)\big)\Big]
  \label{eq:att}
\end{equation}
where $\Vert$ denotes concatenation,
$\mathbf{W}^{k}\in\mathbb{R}^{16\times64}$ and $\mathbf{a}^{k}\in\mathbb{R}^{32}$ are learned parameters for head $k$, and $\mathbf{b}\in\mathbb{R}^{64}$ is a learned
bias. The softmax is evaluated over
$j\in\mathcal{N}(i)\cup\{i\}$, including the self-loop, and the LeakyReLU. The dropout probability in Eq.~\eqref{eq:block} is 0.15; attention-coefficient dropout is disabled.
The attention coefficients allow each vertex to assign different weights to its neighbors based on their learned
embeddings.

\textbf{Descriptor branch.} The 12 descriptors pass through a two-layer encoder (linear $12\rightarrow64$, layer normalization, SiLU, dropout, linear $64\rightarrow64$, SiLU), and the resulting part embedding is copied to every node of the part.

\textbf{Decoder.} Each node's graph embedding and the part embedding are concatenated (128 channels) and mapped by a multilayer perceptron ($128\rightarrow128\rightarrow64\rightarrow3$, with layer normalization and dropout after the first layer and SiLU activations) to predict the standardized displacement components. The predicted standardized displacement components are transformed back to physical units (mm) using the target means and standard deviations fitted on the training data. The full GAT model has about 40.0k trainable parameters.

\subsubsection{Message-passing operators (GCN, GraphSAGE, GAT) and capacity-matched MLP}
\label{Message-passing operators (GCN, GraphSAGE, GAT) and capacity-matched MLP}
To isolate the effect of the message-passing rule, three variants replace only the layer MP in Eq.~\eqref{eq:block}, keeping the encoders, decoder, depth, and training identical: (i) GCN \cite{kipf2017gcn}, with symmetric degree normalization and self-loops; (ii) GraphSAGE \cite{hamilton2017graphsage} with mean aggregation; and (iii) a capacity-matched MLP block that uses no edges, so each node is processed independently. The MLP baseline has approximately as many trainable parameters as the largest graph model, so any advantage of the graph models cannot be attributed to extra capacity.
\subsection{Model Training}
\label{Model Training}
Model performance was estimated with five-fold cross-validation split by part, so that all vertices of a part belong to the same fold. Folds were stratified by each part's area-weighted 95th-percentile displacement magnitude, grouped into five quantile bins. Within each fold, 20\% of the training parts (stratified in the same way) were held out as a validation set for learning-rate scheduling and early stopping, and the remaining parts were used to update the model weights. Therefore, each part appears in the test set exactly once, and all models were trained and evaluated on identical folds with identical settings.

Node features, descriptors, and targets were standardized with means and standard deviations fitted on the training parts of each fold \cite{bishop1995neural}. The loss is a masked, area-weighted, part-balanced smooth L1 loss, a scaled Huber loss \cite{girshick2015fast}, as shown in Eq. (\ref{eq:huber}). 

\begin{equation}  
  \ell_\beta(e)=
  \begin{cases}
    e^2/(2\beta), & |e|<\beta\\
    |e|-\beta/2, & \text{otherwise}
  \end{cases}
  \label{eq:huber}
\end{equation}

with $\beta=0.5$ in standardized units. Area weighting makes the loss approximate a surface-area average, reducing sensitivity to local mesh density. Part balancing gives every part equal total weight regardless of its vertex count.

Models were trained using AdamW with a learning rate of
$10^{-3}$, weight decay of $10^{-5}$, a mini-batch size of
two parts, and gradient-norm clipping at 2.0 for up to
200 epochs. A validation-loss scheduler halved the learning rate with a patience of 15 epochs, down to a minimum of $10^{-6}$. Early stopping was applied after 60 epochs without improvement in validation loss, and the best validation checkpoint was retained. A base random seed of 42 was used. Models were implemented using PyTorch Geometric \cite{fey2019fast} and trained on an Intel(R) Core (TM) Ultra 5 238V (2.10 GHz) CPU.

\subsubsection{Evaluation metrics}
\label{Evaluation Metrics}
All prediction accuracy was evaluated on the valid (unmasked) vertices of the test parts in each cross-validation fold. Predicted displacement components were transformed back to millimeters before evaluation. Each valid vertex was weighted by its assigned surface area divided by the total assigned area of the valid vertices in its part. The weights therefore sum to one within each part, giving every test part equal total weight regardless of its vertex count.

The fold-level metrics were computed over all valid test vertices, separately for each displacement component ($U_x$, $U_y$, and $U_z$) and the displacement magnitude $\lVert\mathbf{u}\rVert$. In the following equations, $y_i$ and $\hat{y}_i$ denote the reference and predicted values, respectively; $w_i$ is the vertex weight; and $\bar{y}_w=\sum_i w_i y_i/\sum_i w_i$ is the weighted-mean reference value.
 
\begin{align}
  R^2&=1-\frac{\sum_i w_i\,(\hat{y}_i-y_i)^2}{\sum_i w_i\,(y_i-\bar{y}_w)^2}, \label{eq:r2}\\
  \text{MAE}&=\frac{\sum_i w_i\,|\hat{y}_i-y_i|}{\sum_i w_i}, \label{eq:mae}\\
  \text{RMSE}&=\sqrt{\frac{\sum_i w_i\,(\hat{y}_i-y_i)^2}{\sum_i w_i}}
   \label{eq:rmse}
\end{align}

Because the vertex weights sum to one within each part, the fold-level MAE and MSE equal the arithmetic means of the corresponding area-weighted per-part values. The fold-level RMSE is the square root of this mean MSE. We report $R^2$ for the three displacement components and the displacement magnitude, together with magnitude MAE and RMSE.

Magnitude-based errors compare
$\lVert\hat{\mathbf{u}}_i\rVert$ with
$\lVert\mathbf{u}_i\rVert$ and therefore do not capture
differences in displacement direction. To assess both magnitude and direction, we also report the Euclidean difference between predicted and reference displacement vectors. Denoting this as vector MAE,
\begin{equation}
  e_i=\lVert\hat{\mathbf{u}}_i-\mathbf{u}_i\rVert,
  \label{eq:endpoint}
\end{equation}
Here, it is averaged using the same area weights and part balancing (vector MAE). Cross-validation results are summarized as the mean $\pm$ standard deviation over the five folds \cite{akbari2022meltpoolnet, xie2021mechanistic}.

The pooled $R^2$ in Eq.~\eqref{eq:r2} reflects displacement variation both within and between parts. To additionally assess prediction accuracy within individual parts, we analyzed the median per-part $R^2$, calculated separately over the valid vertices of each test part using the same area weighting.
As a sensitivity check, unweighted fold-level $R^2$ was also computed using the built-in TorchMetrics implementation. This corresponds to Eq.~\eqref{eq:r2} with $w_i=1$ and an unweighted reference mean. In this variant, parts with more valid vertices contribute more to the score.

To compare parts of different sizes, the magnitude MAE of each test part $p$ was normalized by its enclosed volume $V_p$:
\begin{equation}
  \eta_p=\frac{\text{MAE}_p}{V_p}\;
  \label{eq:relerr}
\end{equation}
Here, $\eta_p$ is summarized by its median and interquartile range over the 117 test parts.

\section{Results}
\label{Results}
This section first explains the accuracy of the transferred labels (Section~\ref{Label-transfer accuracy}) and then compares the four models on the test parts of the five folds (Section~\ref{Model comparison}). Then (Section~\ref{Per-part accuracy}) describes how error varies across parts, (Section~\ref{Descriptor importance}) shows how much each group of global descriptors contributes to the prediction, and (Section~\ref{Masking sensitivity}) explains whether label masking affects the results. All values refer to test parts not used for training, model selection, or scaling in their fold.

\subsection{Label-transfer accuracy}
\label{Label-transfer accuracy}
\begin{figure*}[t]
  \centering
  \includegraphics[width=\textwidth]{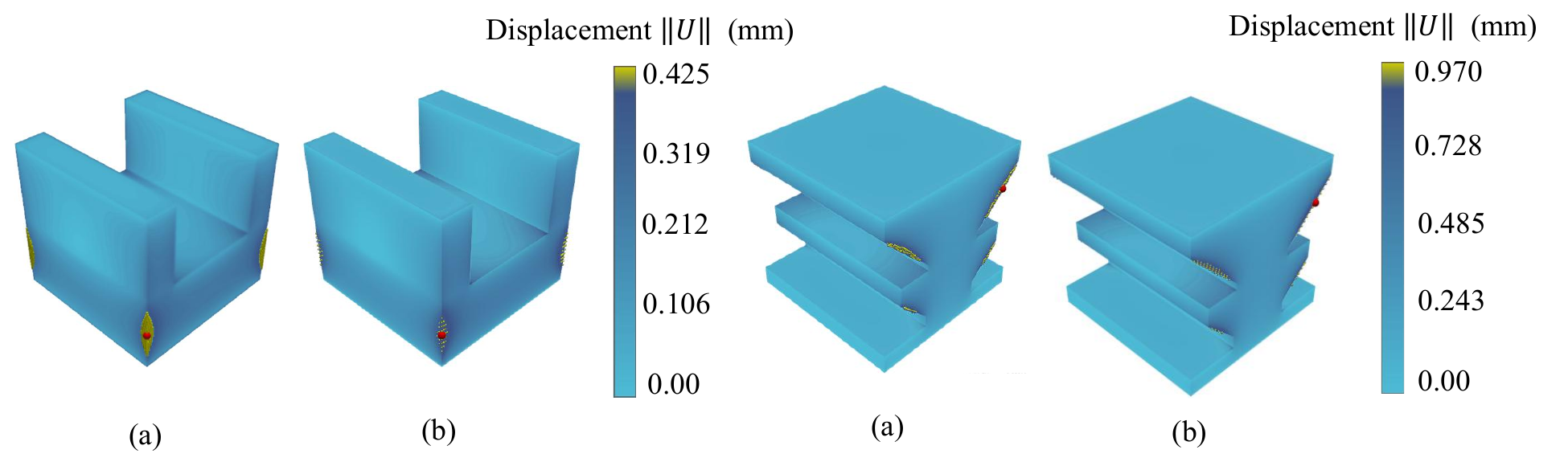}
  \caption{Displacement magnitude $\lVert\mathbf{u}\rVert$ of parts (a)~p1758 and (b)~p1768 before and after label transfer. For each part, the left image shows the surface of the voxel mesh from the process simulation (VTK), and the right image shows the displacements transferred to the densified STL surface using a Gaussian kernel. Yellow dots mark the points above the 99th percentile of $\lVert\mathbf{u}\rVert$ and the red dot marks the peak} 
  \label{fig:label_transfer}
\end{figure*}

Reconstructing held-out simulation surface nodes with the transfer kernel (Section~\ref{Data pre-processing}) yielded an MAE of 0.0023~mm, ranging from 0.0009 to 0.0067~mm across ten randomly selected parts. This is about ten times smaller than the vector MAE of the best model (0.0230~mm), supporting the accuracy of the Gaussian-weighted average for generating training labels. Figure~\ref{fig:label_transfer} compares the displacement magnitude on the VTK surface with the transferred values on the STL surface for two parts. Across the 117 parts, the transfer changed the peak displacement by a median of 0.009~mm (2.2\% on average), about four times less than the peak error of the best model (median 0.034~mm), and the 99th-percentile displacement by a median of 0.008~mm.

\subsection{Model comparison}
\label{Model comparison}
Table~\ref{tab:models} compares the test performance of the four models, averaged over five folds. All three graph models achieve higher mean $R^2$ values than the capacity-matched MLP for the displacement magnitude and each displacement component. GAT achieves the highest magnitude $R^2$ ($0.946\pm0.022$), compared with $0.936\pm0.024$ for GraphSAGE, $0.934\pm0.027$ for GCN, and $0.913\pm0.037$ for the MLP. GAT also achieves the highest mean $R^2$ for $U_x$ and $U_z$, whereas GraphSAGE performs slightly better for $U_y$. The differences in $U_y$ are small, with all four models achieving $R^2$ values between 0.972 and 0.976.

The improvements associated with message passing differ across displacement components. The largest increase in mean $R^2$ occurs in the build-direction component, $U_z$, which rises from 0.903 for the MLP to 0.939 for GAT. In contrast, the high $U_y$ accuracy of the MLP leaves less room for improvement through graph-based aggregation. These results suggest that neighborhood information is particularly beneficial for predicting $U_z$ in this dataset. The unweighted magnitude $R^2$ values are 0.004--0.005 lower than their area-weighted counterparts and preserve the same model ranking: GAT (0.942), GraphSAGE (0.932), GCN (0.929), and MLP (0.909). Thus, the observed ranking is robust to the choice between these two weighting schemes.

\begin{figure}[htbp]
    \centering
    \includegraphics[width=.6\columnwidth]{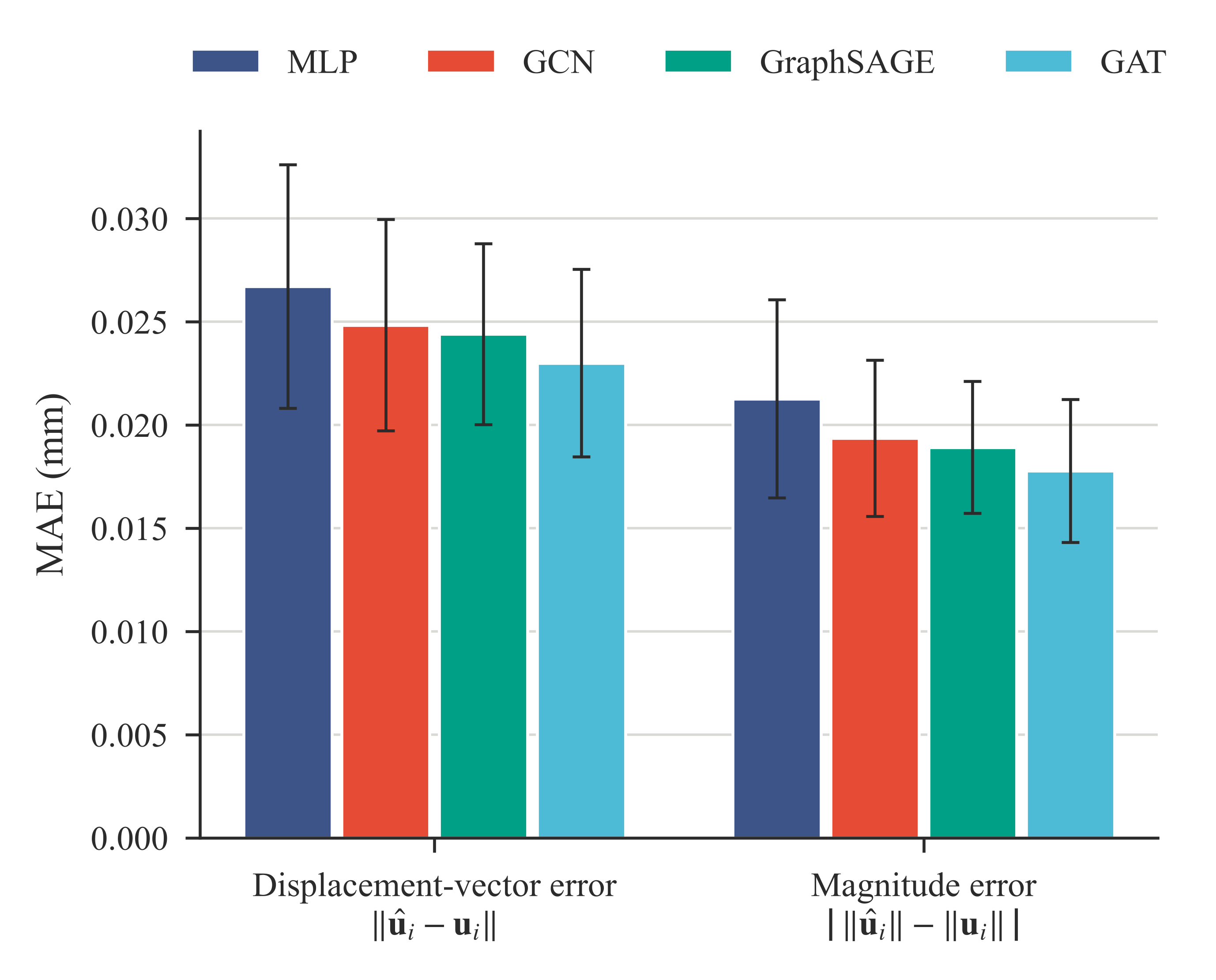}
    \caption{Test displacement vector error and magnitude MAE for the
    four models. Bars show the mean across five folds, and error bars indicate one standard deviation.}
    \label{fig:mae}
\end{figure}

\begin{figure}[htbp]
    \centering
    \includegraphics[width=.6\columnwidth]{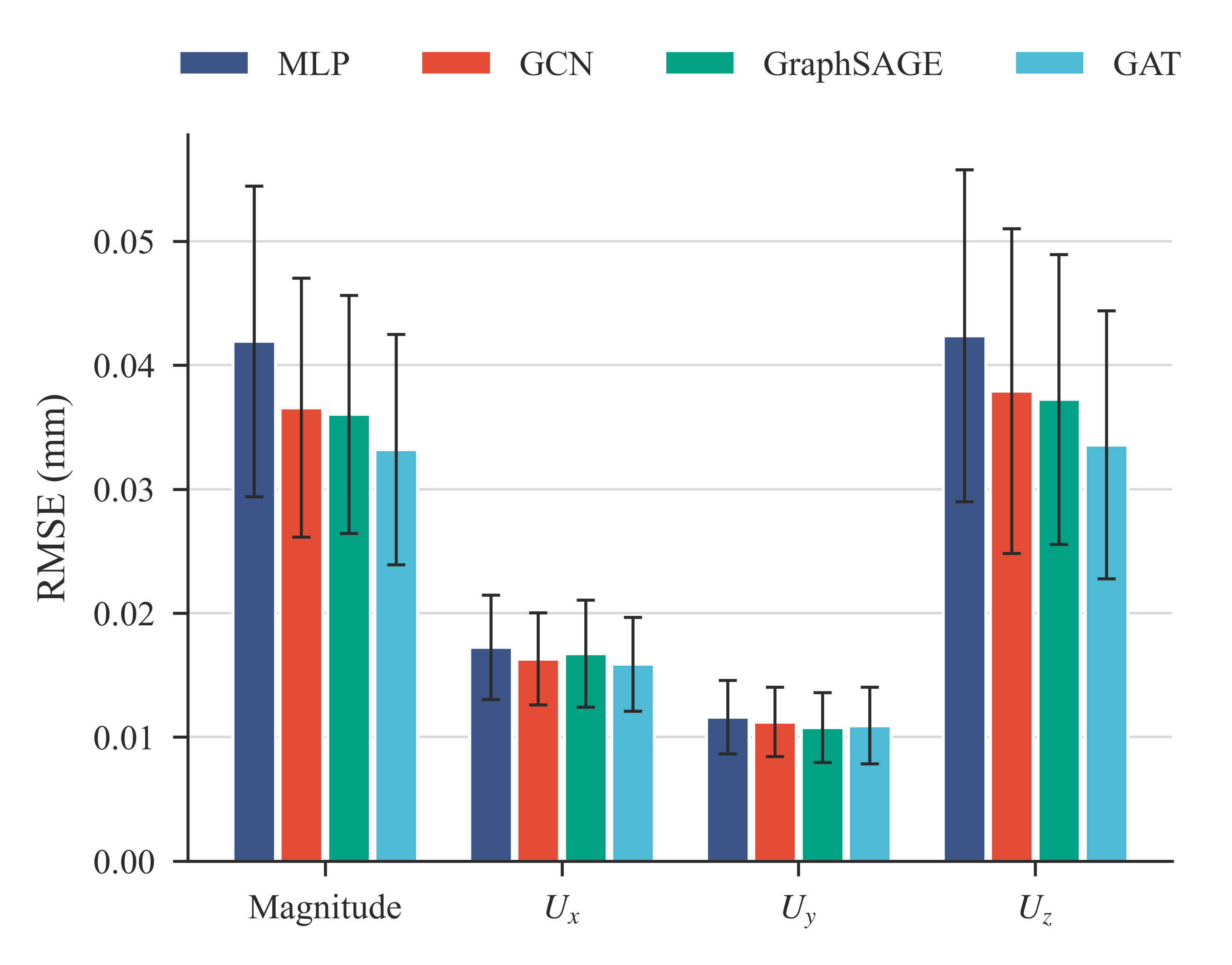}
    \caption{Test RMSE for the displacement magnitude and the
    three displacement components. Bars show the mean across five folds, and error bars indicate one standard deviation.}
    \label{fig:rmse}
\end{figure}

Figures~\ref{fig:mae} and~\ref{fig:rmse} present the test MAE
and RMSE, respectively. GAT achieves the lowest mean magnitude
MAE ($0.0178\pm0.0035$~mm) and magnitude RMSE
($0.0332\pm0.0093$~mm), compared with $0.0213\pm0.0048$~mm
and $0.0419\pm0.0125$~mm for the MLP.
GAT also achieves the lowest displacement vector MAE
($0.0230\pm0.0045$~mm; Table~\ref{tab:models}), indicating
improved prediction of the displacement vectors as well as their
magnitudes. Calculating magnitude MSE within each test fold
and then averaging across the five folds shows that GAT reduces
mean magnitude MSE by 37.8\% relative to the MLP, 17.8\%
relative to GCN, and 14.6\% relative to GraphSAGE.
GAT achieves higher magnitude $R^2$ and lower vector MAE
than each competing model in all five folds.
GCN and GraphSAGE perform similarly, and both outperform
the MLP in every fold on these two metrics, with mean magnitude
MSE reductions of 24.3\% and 27.1\%, respectively.
The consistent improvements across folds support the benefit
of incorporating mesh connectivity through message passing
under the evaluation protocol used in this study.

\begin{figure}[t]
    \centering
    \includegraphics[width=.6\columnwidth]{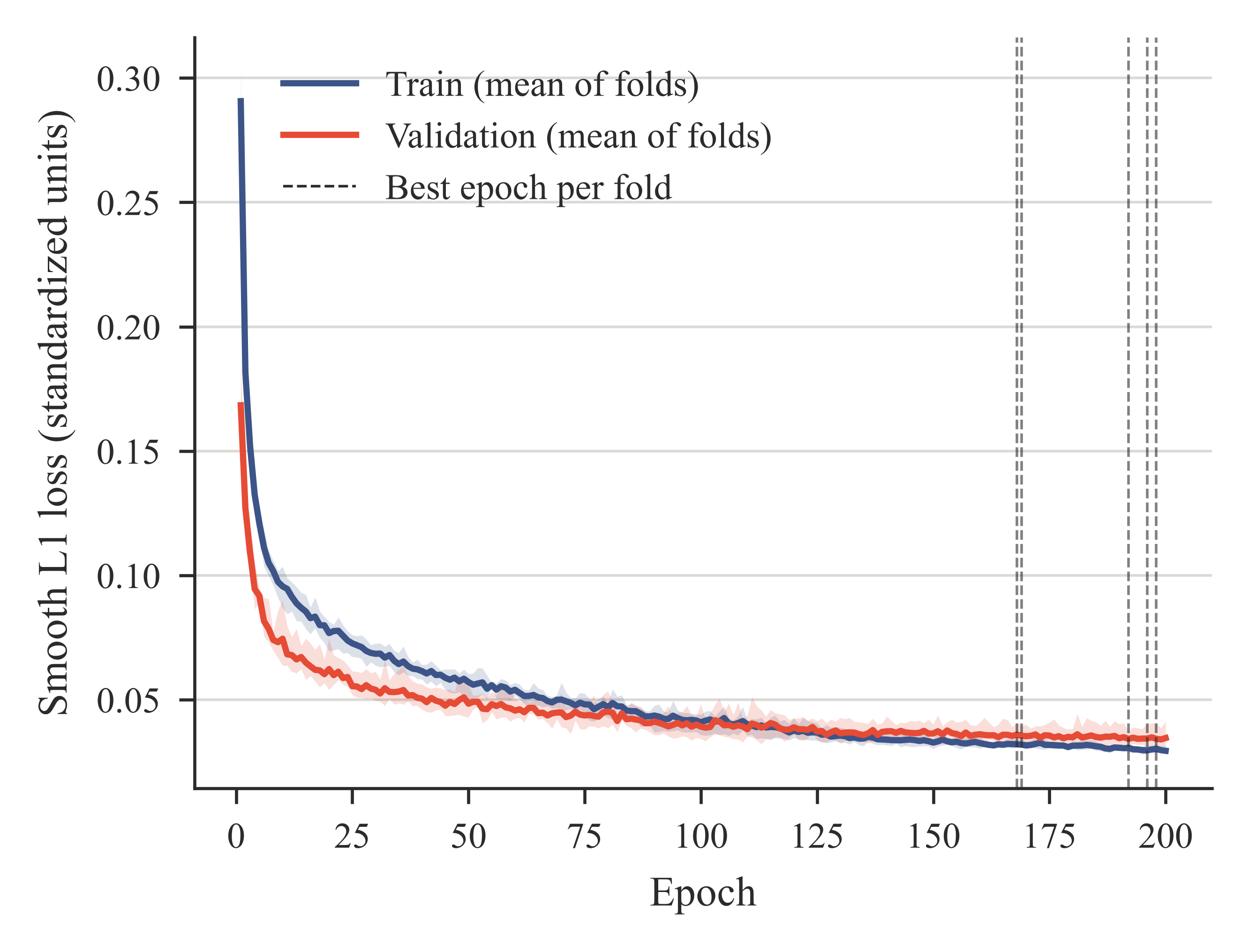}
    \caption{Training and validation loss of the GAT model across epochs for all five folds. The loss is the masked, area-weighted, Smooth L1 loss in standardized displacement
    units.}
    \label{fig:loss}
\end{figure}

The mean training and validation losses of GAT (Fig.~\ref{fig:loss}) decrease rapidly during the initial epochs, followed by more gradual improvement. At later epochs, the validation loss approximately plateaus while the training loss continues to decrease slightly, leaving a small gap. The averaged curves indicate stable convergence and close agreement between the mean training and validation losses.

\begin{table*}[htb]
  \centering
  \caption{Test performance of the four models (mean over the five folds, $\pm$ 1SD where given). Best value per column in bold.}
  \label{tab:models}
  \setlength{\tabcolsep}{3pt}
  \begin{tabular*}{\textwidth}{@{\extracolsep{\fill}}lcccccc@{}}
    \toprule
    & \multicolumn{4}{c}{$R^2$ (area-weighted)}
    & $R^2$ & Displacement vector \\
    \cmidrule(lr){2-5}
    Model & Magnitude & $U_x$ & $U_y$ & $U_z$
    & unweighted & MAE (mm) \\
    \midrule
    MLP
    & $0.913\pm0.037$ & 0.919 & 0.972 & 0.903
    & 0.909 & $0.0267\pm0.0059$ \\
    GCN
    & $0.934\pm0.027$ & 0.928 & 0.974 & 0.922
    & 0.929 & $0.0248\pm0.0051$ \\
    GraphSAGE
    & $0.936\pm0.024$ & 0.924 & \textbf{0.976} & 0.925
    & 0.932 & $0.0244\pm0.0044$ \\
    GAT
    & $\mathbf{0.946\pm0.022}$ & \textbf{0.931} & 0.975
    & \textbf{0.939} & \textbf{0.942}
    & $\mathbf{0.0230\pm0.0045}$ \\
    \bottomrule
  \end{tabular*}
\end{table*}

\subsection{Per-part accuracy}
\label{Per-part accuracy}
\begin{figure}[t]
  \centering
  \includegraphics[width=.6\columnwidth]{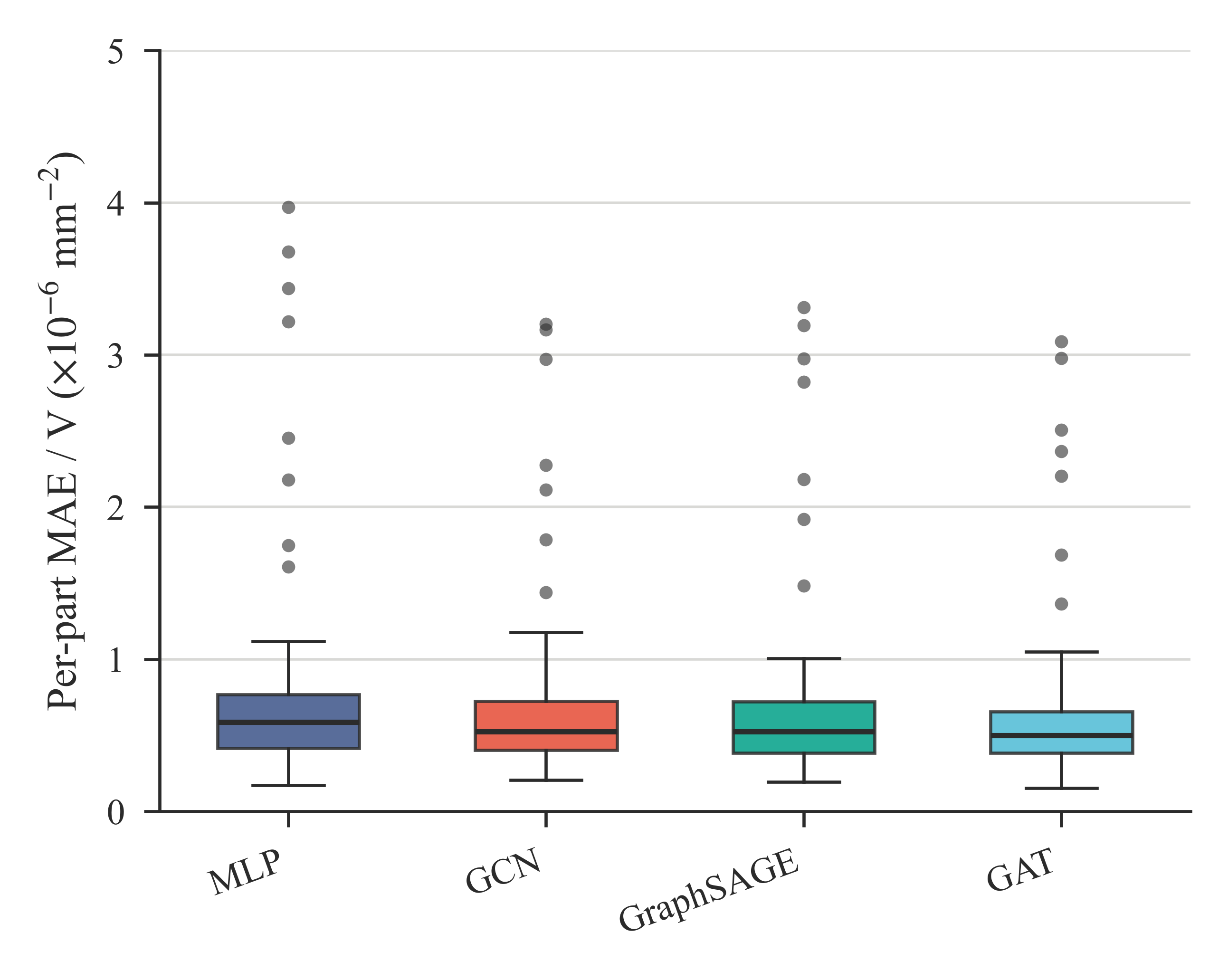}
  \caption{Per-part relative error $\eta_p$ (magnitude MAE divided by part volume) on the 117 test parts for each model. Boxes show the interquartile range, the line inside each box marks the median. The axis is limited to $5\times10^{-6}$~mm$^{-2}$; the two small parts, p2076 and p2080, lie above this range for all models.}
  \label{fig:relerr}
\end{figure}

GAT has a median magnitude $R^2$ of 0.963 and a median relative error $\eta_p$ (Eq.~\eqref{eq:relerr}) of $0.50\times10^{-6}$~mm$^{-2}$ (interquartile range 0.38--0.66$\times10^{-6}$~mm$^{-2}$), compared with $0.59\times10^{-6}$~mm$^{-2}$ (0.42--0.77$\times10^{-6}$~mm$^{-2}$) for the MLP (Fig.~\ref{fig:relerr}). Comparing the models part by part, GAT has lower error than MLP on 100 of the 117 parts, than GCN on 92, and than GraphSAGE on 80. The differences between models are therefore small compared with the differences between parts, but consistent across parts.

The largest relative errors are concentrated in a similar group of parts across all four models, with eight parts consistently appearing among the nine highest-error cases. The parts are- p1756, p1784, p1786, p1878, p1808, p1870, the two small parts (p2076, p2080); the MLP has one additional outlier p1964. For GAT, 90\% of the parts have $\eta_p<0.87\times10^{-6}$~mm$^{-2}$, while p2076 and p2080 reach 7.5 and $12.6\times10^{-6}$~mm$^{-2}$ (above the range shown in Fig.~\ref{fig:relerr}). Their MAE (0.042 and 0.045~mm) is among the ten largest of all parts, and dividing by their small volumes amplifies it further. Parts p1756 and p1878 have two of the largest absolute errors (MAE 0.092 and 0.088~mm), and p2080 has the lowest per-part $R^2$ ($-1.72$), as it combines the smallest peak displacement in the dataset (0.12~mm) with the largest fraction of masked labels (12\%). Because the same parts are outliers for all four models, their larger errors appear to stem from the parts rather than from a particular architecture. For the parts with higher erros, the most likely cause is that these configurations are underrepresented in the training data.

\subsection{Descriptor importance}
\label{Descriptor importance}
Table~\ref{tab:ablation} shows the effect of removing each descriptor group from GAT. Removing the mass-distribution descriptors (fill ratio and centroid height) causes the largest loss: magnitude $R^2$ drops from 0.946 to 0.900, the magnitude MAE rises by 40.4\%, in all five folds ($p<0.001$). Removing the base-contact descriptors lowers $R^2$ to 0.920 and raises the MAE by 23.5\% ($p=0.004$). Removing the size descriptors increases MAE by 7.63\% in all five folds ($p=0.002$), but the change in $R^2$ is not significant ($p=0.075$). Removing the surface-to-volume ratio has no significant effect, consistent with its information being contained in the surface area and volume, which remain in the size group.

\begin{table}[t]
  \centering
  \caption{Effect of removing one descriptor group from GAT (test, mean over the five folds; $p$: paired $t$-test on magnitude $R^2$).}
  \label{tab:ablation}
  \setlength{\tabcolsep}{3pt}
  \begin{tabular*}{\linewidth}{@{\extracolsep{\fill}}lcccc@{}}
    \toprule
    Removed group & Mag.\ $R^2$ & $\Delta$MAE\% & $p$ \\
    \midrule
    None (full model)   & 0.946 & --              & --       \\
    Mass distribution   & 0.900 & $+40.4\%$     & $<0.001$ \\
    Base contact and overhang        & 0.920 & $+23.5\%$    & 0.004    \\
    Size                & 0.938 & $+7.63\%$    & 0.075    \\
    Surface to volume ratio      & 0.943 & $-0.56\%$    & 0.35     \\
    \bottomrule
  \end{tabular*}
\end{table}

\subsection{Masking sensitivity}
\label{Masking sensitivity}
Masking excludes vertices with unreliable transferred labels from the loss and from all metrics. To check that it does not impact the reported accuracy, GAT was retrained on the same folds with all vertices treated as valid, so that the 592 previously masked vertices (0.04\%) entered both the training and the evaluation. The magnitude $R^2$ changed from 0.9458 to 0.9448 and the displacement magnitude MAE from 0.01777 to 0.01784~mm, while the per-part MAE did not differ significantly. The only part with a noticeable change was p2080, which had the largest masked fraction. The masking therefore removes unreliable labels, making the model efficient without affecting the reported accuracy.

\section{Discussion}
\label{Discussion}
The results show that the full surface displacement field of LPBF parts can be predicted from the STL geometry, and that both the graph structure and the part-level descriptors contribute to this accuracy. This section discusses why the models differ, what the descriptor ablation indicates,
and where the approach is least accurate.

\subsection{Why message passing and attention help}
The MLP model receives the same node features and descriptors as the graph models but processes each vertex independently. It therefore knows where a vertex lies in the part and how its surface is oriented, but not how the surrounding surface is connected. Distortion, however, depends on the geometry and structural connectivity of the part--how much material lies above or below a region, attachment to the build plate, material distribution, and where the geometry changes abruptly \cite{gao2024experimental,Chandran2025}. Thus, message passing enables each vertex to incorporate information from neighboring vertices, contributing to improved predictive performance over the MLP baseline across all five folds (Section~\ref{Model comparison}).

The largest gain occurs in the build direction ($U_z$ $R^2$ increases from 0.903 to 0.939 for GAT), while all models predict $U_y$ almost equally well. A plausible explanation is that build-direction distortion depends most strongly on how material is stacked along the part height, which is exactly the vertical connectivity the mesh graph provides. Two message-passing layers cover roughly two mesh edges (about 1.7~mm), so this information is local and the part-level context comes from the descriptor branch.

\subsection{Why attention outperforms fixed aggregation}
\label{Why attention outperforms fixed aggregation}

GCN, GraphSAGE, and GAT primarily differ in how they aggregate information from neighboring vertices. GCN combines a vertex's features with those of its neighbors using degree-normalized weights, while GraphSAGE uses a mean aggregator that assigns equal weights to neighboring vertices. In contrast, GAT employs attention mechanisms to learn the relative importance of neighboring vertices during feature aggregation. Attention can allow a vertex to favor the neighbors most relevant to it, for example, those on the same face, whereas a fixed or equal weighting mixes them. This is consistent with GAT outperforming both GCN and GraphSAGE on all five folds, while the two fixed-weighting operators (GCN and GraphSAGE) perform similarly.

\subsection{The role of part-level descriptors}
\label{The role of part-level descriptors}
Because the nodal features are normalized by part dimensions, they primarily describe relative geometric characteristics, while the part-level descriptors provide explicit information about global geometry and scale. The ablation (Table~\ref{tab:ablation}) study shows that the mass-distribution and base-contact-and-overhang groups carry most of this information. This partly corresponds with the study conducted by Bushra et al.(2023), where the features describing material distribution across layers and the shape at the connection to the build plate, were the most important while predicting peak distortion value for similar type of parts \cite{bushra2023}.The fill ratio and centroid height describe how material is distributed over the build height, which governs how much residual stress can accumulate above a given layer \cite{bushra2023}. The base-contact and overhang descriptors describe how the part is anchored to the build plate and how much of its surface faces downward, which determines how the plate restrains it \cite{gao2024experimental}. Removing the size group mainly increased the displacement vector MAE, which is expected because absolute size sets the magnitude of the displacements. In contrast, the geometric pattern is already captured by the dimensionless node features. The surface-to-volume ratio had no measurable effect, as it can be derived from the surface area and volume that remain in the size group. Because several descriptors are correlated, the remaining groups can partly compensate when one group is removed; the ablation therefore measures the information unique to each group, and a small effect does not imply that the information is unimportant, only that it is also available from other descriptors.

\subsection{Use in design-for-AM workflows}
\label{Use in design-for-AM workflows}
Once trained, the model predicts the surface displacement field of an unseen part in a single forward pass with a median time of 32~ms on a laptop CPU (Intel Core Ultra 5 238V), excluding graph construction. For comparison, thermal strain simulations of parts from the same design family required 1 to 8~h for smaller parts and up to 90~h for larger ones \cite{bushra2023}, so the prediction is several orders of magnitude faster. This speed-up applies after a one-time cost of generating the simulated training data and training the model (about 6~h) with a displacement vector MAE of 0.023~mm. This makes it suitable for screening and comparing design variants early in the design process, for example to identify geometries that are likely to distort.

\section{Conclusion}
\label{Conclusion}
This study develops a two-branch graph attention framework that predicts the full surface distortion field of LPBF parts directly from their STL derived graphs. Simulated displacements were transferred from the VTK mesh of the process simulation to the densified STL surface with a Gaussian kernel-weighted average, and a two-branch network combined message passing over the surface mesh with twelve part-level geometric descriptors.

The results show that the attention-weight mechanism over the mesh improves prediction at equal model capacity. All graph models outperform a capacity-matched node-wise MLP in every fold, with the largest gain in the build direction. Among the three graph model types tested, graph attention performed surface displacement prediction most accurately. GAT reaches a magnitude $R^2$ of 0.946 and a vector MAE of 0.023~mm, with a magnitude MSE 38\% lower than the MLP and 15--18\% lower than GCN and GraphSAGE and has the lowest error of all models in all five folds.

Additionally, an ablation study involving dropping different groups of descriptors shows that part-level context is essential. Descriptors of mass distribution and base contact and overhang contribute most to the accuracy, whereas the surface-to-volume ratio is redundant. Another important observation is that errors concentrate on configurations that are rare in the training data, in particular the comparatively smaller parts, and this pattern is the same across all models.

The present results are limited to one material and one set of process parameters, to labels obtained from simulation rather than measurement, and to 117 mostly prismatic parts with insufficient variation. Future work will complement this study by incorporating an extended dataset with freeform and thin-walled geometries with more varied support conditions, process parameters, and materials. Despite these limitations, this study demonstrates that graph neural networks operating directly on STL derived graphs can capture the full surface distortion field of LPBF parts with high accuracy. The STL-to-label pipeline and the controlled comparison of graph based operators establish a baseline on which models covering more geometries, materials, and process conditions can be built. The proposed framework thus offers a fast screening tool for identifying distortion-prone designs early in the design process and provides a baseline for future geometry-based distortion prediction. 

\bibliographystyle{plainnat}
\bibliography{my_collection}

\end{document}